\documentclass[letterpaper, 10 pt, conference]{ieeeconf}  

\usepackage{times}
\usepackage{fix-cm}
\usepackage{etex}

\usepackage{dblfloatfix}

\usepackage{nag}

\makeatletter
\@ifpackageloaded{xcolor}{}{%
\usepackage[table,x11names,dvipsnames,svgnames]{xcolor}%
}
\makeatother

\usepackage{colortbl}

\usepackage{graphicx}
\usepackage{wrapfig}

\definecolorset{RGB}{lyft}{}{Red,194,39,36;Sunset,202,53,33;Orange,205,68,20;Amber,200,117,42;Yellow,242,169,52;Citron,186,188,44;Lime,112,159,33;Green,56,139,31;Mint,45,118,56;Teal,52,133,135;Cyan,60,132,202;Blue,55,94,248;Indigo,64,13,247;Purple,115,42,248;Pink,176,25,145;Rose,176,32,75}

\usepackage{cite}

\usepackage{microtype}

\usepackage{array}
\usepackage{multirow}
\usepackage{booktabs}
\usepackage{makecell} 

\ifcsname labelindent\endcsname
\let\labelindent\relax
\fi
\usepackage[inline]{enumitem}

\usepackage{subfig}

\newenvironment{lenumerate}[2][]
{\begin{enumerate}[label=(#2\arabic*),leftmargin=0.2in,itemindent=0.15in,#1]}
{\end{enumerate}}

\setlist*[enumerate,1]{label={\itshape\arabic*)}}

\makeatletter
\newcommand{\paragraphswithstop}{%
\let\copyparagraph\paragraph%
\renewcommand\paragraph[1]{\copyparagraph{##1.}}%
}
\makeatother

\usepackage[framemethod=tikz]{mdframed}

\usepackage{suffix}

\usepackage{environ}

\makeatletter
\newsavebox{\boxifnotempty}
\newcommand{\displayifnotempty}[3]{\sbox\boxifnotempty{#2}\setbox0=\hbox{\usebox{\boxifnotempty}\unskip}%
\ifdim\wd0=0pt
\else
 #1\usebox{\boxifnotempty}#3%
\fi%
}
\newcommand{\ifempty}[2]{\setbox0=\hbox{#1\unskip}%
\ifdim\wd0=0pt%
 #2%
\fi%
}
\newcommand{\ifnotempty}[2]{\setbox0=\hbox{#1\unskip}%
\ifdim\wd0>0pt%
 #2%
\fi%
}
\makeatother

\usepackage{algpseudocode}
\usepackage{algorithm}

\usepackage{scrlfile}

\makeatletter
\newcommand*\newstoreddef[1]{
  \BeforeClosingMainAux{%
    \immediate\write\@auxout{%
      \string\restoredef{#1}{\csname #1\endcsname}%
    }%
  }%
}
\newcommand*{\restoredef}[2]{
  \expandafter\gdef\csname stored@#1\endcsname{#2}%
}
\newcommand*{\storeddef}[1]{
  \@ifundefined{stored@#1}{0}{\csname stored@#1\endcsname}%
}
\makeatother

\usepackage{pageslts}
\NewEnviron{tee}{\BODY\typeout{Marker Tee [start] ^^J \BODY ^^JMaker Tee [end]}}

\input{math}
\newcommand{\real}[1]{\mathbb{R}^{#1}{}}

\DeclarePairedDelimiter{\norm}{\lVert}{\rVert}

\newcommand{\vct}[1]{\mathbf{#1}}

\DeclareMathOperator{\rank}{rank}

\DeclareMathOperator{\trace}{tr}

\newcommand{\intersect}{\cap}

\newcommand{\subjectto}{\textrm{subject to }}

\providecommand{\vb}{\vct{b}}

\providecommand{\vp}{\vct{p}}

\providecommand{\vu}{\vct{u}}

\providecommand{\mA}{\vct{A}}

\providecommand{\mC}{\vct{C}}

\providecommand{\mP}{\vct{P}}

\providecommand{\mR}{\vct{R}}

\providecommand{\mT}{\vct{T}}
\providecommand{\mU}{\vct{U}}

\providecommand{\mY}{\vct{Y}}

\providecommand{\cE}{\mathcal{E}}

\providecommand{\cG}{\mathcal{G}}

\providecommand{\cR}{\mathcal{R}}
\providecommand{\cS}{\mathcal{S}}

\providecommand{\cU}{\mathcal{U}}
\providecommand{\cV}{\mathcal{V}}
\providecommand{\cW}{\mathcal{W}}

\providecommand{\cY}{\mathcal{Y}}

\def\IR{\mathbb R}			
\def\IS{\mathbb S}
\newcommand{\SO}[1]{\mathbf{SO}(#1)}

\usepackage{units}

\newcommand{\newcolorlabel}[2]{%
  \expandafter\newcommand\csname #1\endcsname[1]{%
    \colorbox{#2}{\color{white}\textsf{\textbf{##1}}}}%
}

\newcommand{\newcommenter}[2]{%
  \expandafter\newcommand\csname #1\endcsname[1]{%
    \fcolorbox{#2}{#2}{\color{white}\textsf{\textbf{#1}}}
    {\color{#2}##1}}%
  \expandafter\newcommand\csname at#1\endcsname{%
    \fcolorbox{#2}{#2}{\color{white}\textsf{\textbf{@#1}}}
    {\color{#2}}}%
  \expandafter\newcommand\csname #1hl\endcsname[2]{%
    \colorbox{#2}{\color{white}\textsf{\textbf{#1}}}\sethlcolor{Azure2}\hl{##2}~%
    \expandafter\ifx\csname commentarrow\endcsname\relax$\leftarrow$\else \commentarrow[#2]\fi~%
    {\color{#2}##1}}%
  \expandafter\newcommand\csname #1st\endcsname[2]{%
    \colorbox{#2}{\color{white}\textsf{\textbf{#1}}}\sout{##2}~%
    \expandafter\ifx\csname commentarrow\endcsname\relax$\leftarrow$\else \commentarrow[#2]\fi~%
    {\color{#2}##1}}%
}
\newcommenter{TODO}{DodgerBlue1}
\newcommenter{rtron}{Green3}

\usepackage{comment}

\usepackage{pdfcomment}

\usepackage{soul}

\usepackage[normalem]{ulem}

\usepackage{csquotes}

\usepackage{tikz}
\usetikzlibrary{calc}
\usetikzlibrary{matrix}
\usetikzlibrary{chains}
\usetikzlibrary{shapes.geometric}
\usetikzlibrary{arrows.meta}
\usetikzlibrary{decorations.pathreplacing}
\usetikzlibrary{backgrounds}

\tikzset{
  dim above/.style={to path={\pgfextra{
        \pgfinterruptpath
        \draw[>=latex,|->|] let
        \p1=($(\tikztostart)!1.5em!90:(\tikztotarget)$),
        \p2=($(\tikztotarget)!1.5em!-90:(\tikztostart)$)
        in(\p1) -- (\p2) node[pos=.5,sloped,above]{#1};
        \endpgfinterruptpath
      }
    }
  },
  dim double above/.style={to path={\pgfextra{
        \pgfinterruptpath
        \draw[>=latex,|->|] let
        \p1=($(\tikztostart)!3em!90:(\tikztotarget)$),
        \p2=($(\tikztotarget)!3em!-90:(\tikztostart)$)
        in(\p1) -- (\p2) node[pos=.5,sloped,above]{#1};
        \endpgfinterruptpath
      }
    }
  },
  dim below/.style={to path={\pgfextra{
        \pgfinterruptpath
        \draw[>=latex,|->|] let 
        \p1=($(\tikztostart)!-1em!-90:(\tikztotarget)$),
        \p2=($(\tikztotarget)!-1em!90:(\tikztostart)$)
        in (\p1) -- (\p2) node[pos=.5,sloped,below]{#1};
        \endpgfinterruptpath
      }
    }
  },
}

\tikzset{
    right angle quadrant/.code={
        \pgfmathsetmacro\quadranta{{1,1,-1,-1}[#1-1]}     
        \pgfmathsetmacro\quadrantb{{1,-1,-1,1}[#1-1]}},
    right angle quadrant=1, 
    right angle length/.code={\def\rightanglelength{#1}},   
    right angle length=2ex, 
    right angle symbol/.style n args={3}{
        insert path={
            let \p0 = ($(#1)!(#3)!(#2)$) in     
                let \p1 = ($(\p0)!\quadranta*\rightanglelength!(#3)$), 
                \p2 = ($(\p0)!\quadrantb*\rightanglelength!(#2)$) in 
                let \p3 = ($(\p1)+(\p2)-(\p0)$) in  
            (\p1) -- (\p3) -- (\p2)
        }
    }
}

\newcommand{\pgfextractangle}[3]{%
    \pgfmathanglebetweenpoints{\pgfpointanchor{#2}{center}}
                              {\pgfpointanchor{#3}{center}}
    \global\let#1\pgfmathresult  
}

\usetikzlibrary{shapes.arrows}
\newcommand{\commentarrow}[1][Azure4]{\tikz[baseline=-3pt]{\node[shape border uses incircle, fill=#1,rotate=180,single arrow, inner sep=1pt, minimum size=6pt, single arrow head extend=2pt]{};}}

\tikzset{ax/.style={-latex,line width=2pt}}

\tikzset{camera/.style={fill=Sienna1,fill opacity=0.5},%
image plane/.style={draw=RoyalBlue3,line width=2pt}}

\usepackage{bm}

\newcommand{\Mat}[1]{\begin{bmatrix}#1\end{bmatrix}}	
\def\trans{\mathrm{T}}

\newcommand{\aleq}[1]{\begin{align}#1\end{align}} 				

\def\IR{\mathbb R}					

\newlength{\figwidth}
\def\autoref#1{\Fref{#1}}
\makeatletter
\def\mkfancyprefix#1#2{%
  \@namedef{fancyref#1labelprefix}{#1}%
  \begingroup\def\x{\endgroup\frefformat{vario}}%
    \expandafter\x\csname fancyref#1labelprefix\endcsname
      {\MakeLowercase{#2}\fancyrefdefaultspacing##1}%
  \begingroup\def\x{\endgroup\Frefformat{vario}}%
    \expandafter\x\csname fancyref#1labelprefix\endcsname
      {#2\fancyrefdefaultspacing##1}%
			\begingroup\def\x{\endgroup\frefformat{plain}}%
    \expandafter\x\csname fancyref#1labelprefix\endcsname
      {\MakeLowercase{#2}\fancyrefdefaultspacing##1}%
  \begingroup\def\x{\endgroup\Frefformat{plain}}%
    \expandafter\x\csname fancyref#1labelprefix\endcsname
      {#2\fancyrefdefaultspacing##1}%
}
\makeatother

\def\IR{\mathbb R}			
\def\IS{\mathbb S}

\def\vec{\mathrm{vec}}

\def\ave{\cE}

\def\trans{{\scriptstyle\ensuremath{\mathsf{T}}}}

\newcommenter{skipped}{white!80!red}
\usepackage{multicol}

\IEEEoverridecommandlockouts                              

\usetikzlibrary{arrows}
\usetikzlibrary{shapes}
\newcommand{\mymk}[1]{%
  \tikz[baseline=(char.base)]\node[anchor=south west, draw,rectangle, rounded corners, inner sep=1.3pt, minimum size=1.5mm,
    text height=2mm](char){\ensuremath{#1}} ;}

\newcommand*\circled[1]{\raisebox{.5pt}{\textcircled{\raisebox{-.9pt} {#1}}}}

\title{\LARGE \bf
 Inverse Kinematics with Vision-Based Constraints
}

\author{Liangting Wu and Roberto Tron
  \thanks{The authors are with Department of Mechanical Engineering, Boston University, 110 Cummington Mall, Boston, MA 02215, USA.
    Emails: {\tt\small tomwu@bu.edu, tron@bu.edu}. 
    The authors gratefully acknowledge the support by NSF award FRR-2212051.
  }
}

\begin{document}

\maketitle
\thispagestyle{empty}
\pagestyle{empty}

\begin{abstract}
This paper introduces the Visual Inverse Kinematics problem (VIK) to fill the gap between robot Inverse Kinematics (IK) and visual servo control. Different from the IK problem, the VIK problem seeks to find robot configurations subject to vision-based constraints, in addition to kinematic constraints. In this work, we develop a formulation of the VIK problem with a Field of View (FoV) constraint, enforcing the visibility of an object from a camera on the robot. Our proposed solution is based on the idea of adding a virtual kinematic chain connecting the physical robot and the object; the FoV constraint is then equivalent to a joint angle kinematic constraint. Along the way, we introduce multiple vision-based cost functions to fulfill different objectives. We solve this formulation of the VIK problem using a method that involves a semidefinite program (SDP) constraint followed by a rank minimization algorithm. The performance of this method for solving the VIK problem is validated through simulations.

\end{abstract}


\section{Introduction}
In robotics, the Inverse Kinematics (IK) problem is the problem of finding the values of joint configurations given a desired end-effector pose. Depending on the robot structure there may exists uncertain number of solutions. Solving the IK problem has been researched extensively over decades, producing numerous approaches classified as analytical \cite{lee1988displacement,raghavan1993inverse} and numerical solutions, e.g., \cite{kenwright2012inverse,Naour2019kinematics,dai2019global}. See \cite{aristidou2018inverse} for a review. This paper investigates a variant of the IK problem, which can be introduced using the following scenario.

\textbf{\textit{Problem motivation.}} Imagine a robot manipulator that needs to inspect an object using a camera attached to its end-effector.
For example, a manipulator makes inspections of a product as part of a manufacturing procedure. Or, for another example, an under water vehicle relies on close-up vision signals from a camera on its manipulator while performing operations in a low visibility environment.
The position of the object is assumed to be generally known (e.g., through a simultaneous localization and mapping (SLAM) algorithm). We would like to find a joint configuration such that the object remains in the camera field of view (FoV) while achieving some objectives, such as keeping the camera upright or viewing the object from a preferred angle. We refer such problem as the Visual Inverse Kinematics (VIK) problem, which is different from the Inverse Kinematics problem because the end-effector pose is not explicitly provided; instead, the feasible set is determined by the intersection between the set of the kinematic constraints of the IK problem and the set of configurations that make the object visible to the camera.

\textbf{\textit{Challenges.}} In addition to the difficulty brought by the kinematic constraints, solving the VIK problem is challenging for the following reasons: \begin{enumerate}
    \item Multiple camera poses that satisfy the field of view constraint may exist. However, due to the kinematic constraints of the robot, not all of the poses are kinematically feasible;\label{itm:challenge1}
    \item The pose of the frustum formed by the camera field of view is a nonlinear function of the joint configuration. Hence, restricting the object in this frustum yields a nonlinear constraint.\label{itm:challenge2} 
\end{enumerate}
A naive approach to the problem would decouple the problem into two: first find a pose of the camera frame feasible to the field of view constraint, and then attempt to solve for the inverse kinematics problem using this pose. However, this approach might fail due to reason \ref{itm:challenge1} alone. Another approach would be to incorporate the field of view constraint in the kinematics problem of the robot and solve it numerically. However, as mentioned in reason \ref{itm:challenge2}, the additional nonlinearity makes regular IK solvers prone to fail due to local minima. 

\textit{\textbf{Applications.}} A related area of research is visual servoing, which is the application of vision data for the feedback control of a robot \cite{chaumette2006visual}. Visual servoing has been applied to different robotic scenarios such as tracking and positioning of Unmanned Aerial Vehicles (UAVs) \cite{sheckells2016optimal,penin2017vision,zheng2019toward}, ground robots \cite{Wei:IJISTA05,cherubini2011visual,wang2009hybrid}, manipulators \cite{kragic2002survey,al2022robotic,dursun2023maintaining}, and etc.

Because the control relies on vision data, it is vital to maintain the target in the Field of View. There have been different approaches to ensure the visibility constraint in visual servoing. To name a few, these include formulating the trajectory planning with visibility constraints as an optimization problem \cite{sheckells2016optimal, penin2017vision,potena2017effective,falanga2018pampc}; using control barrier functions \cite{zheng2019toward,huang2021linear,dursun2023maintaining}; controllers based on machine learning \cite{wang2009hybrid, fu2023deep}.

In this paper, we define and propose a way to solve the visual inverse kinematics problem. Differently from the aforementioned visual-servoing techniques, the VIK problem seeks to find a configuration of the robot that satisfies the visibility and kinematic constraints. Such configuration could be used in visual servoing as an initialization or target configuration (similarly to how IK is used for joint-based control). 

\textit{\textbf{Method summary.}} We formulate the kinematic constraints of the robot using the method introduced in \cite{wu2023cdc} for the VIK problem; this method uses a semidifinite programming (SDP) relaxation followed by a rank minimization technique on fixed-trace matrices. This method, requires to solve only a series of convex problems, and has local convergence guarantees. While the original paper \cite{wu2023cdc} considers only kinematic constraints such as joint axis and angle limits, an expanded algorithm adds prismatic joints in \cite{wu2024tac_archive}. In this paper, we add the visibility constraints as a series of \emph{virtual} prismatic joints. The visibility constraint can then be relaxed to an SDP constraint, and included together with the kinematic constraints to form the feasible set of the VIK problem. Along the way, we propose different vision-based costs to fulfill various objectives (e.g., matching feature positions with respect to an image taken in advance). 

\section{Kinematic constraints Using Rank-1 Semidefinite Matrices}
This section reviews previous work \cite{wu2023cdc} and \cite{wu2024tac_archive} on how to model the kinematic constraints as the intersection of positive semidefinite matrices and rank-1 matrices. We start with a general formulation of the inverse kinematics problem which includes revolute and prismatic joints.
\subsection{Revolute joints}
We parameterize the robot kinematic chain using a graph $\cG=(\cV,\cE)$, where $\cV$ refers to the links and $\cE$ refers to the joints. We denote $\{\mR_i\in\SO{3},\mT_i\in\IR^3\}$ as the set of rigid body transformations from a reference frame attached to the link $i$ to the world reference frame $\cW$. The translation from frame $i$ to a neighboring frame $j$, ${}^{i}\mT_{j}$ is known, fixed, and related to the translations $\{\mT_i\}$ and the rotation $\mR_i$ by
\aleq{
  \mT_j-\mT_i = \mR_i{}^i\mT_{j}.\label{eq:translation relation}
}
Thus we can parameterize the inverse kinematics problem as a function of the rotations matrices
\begin{equation}
    \hat{\mR}=\Mat{\mR_1 &\mR_2 &\dots &\mR_i &\dots &\mR_{n}}.
\end{equation}
We denote a subset of the link indexes, $\cV_r\subset \cV$ such that each associated rotation $\mR_i,i\in\cV_r$ is unknown, and the corresponding matrix $\mR$ as a truncated $\hat{\mR}$ with only unknown rotations. We denote $n_r$ as the number of unknown rotations.

As discussed in \cite{wu2023cdc}, the kinematic constraints of the inverse kinematics problem with revolute joints only can be defined with the set $\cU$ of all the vectors $\vu= \vec(\mR)\in\IR^{9n_r}$ such that
 \begin{subequations}\label{eq:def ik feas set}
    \begin{align}
      &\mA_{\textrm{axis}}\vu = \vb_{\textrm{axis}},\label{eq:ik axis}\\
      &\mA_{\textrm{angle}}\vu \leq \vb_{\textrm{angle}},\label{eq:ik angle}\\
      &\mR_{i}\in\SO{3},\forall i\in\cV_r,\label{eq:ik manifold}
    \end{align}
  \end{subequations}
  where \eqref{eq:ik axis} and \eqref{eq:ik angle} are linear constraint ensuring that the links sharing a revolute joint have common axis and joint angle limit, respectively.

\begin{remark}
    The constraints in \eqref{eq:def ik feas set} can be applied to spherical joints by removing the common axis constraint \eqref{eq:ik axis}. In this case the angle limit constraint \eqref{eq:ik angle} restricts a spatial conic limit of the two links.
\end{remark}

The rotations matrices introduce the nonlinear constraints \eqref{eq:ik manifold}. To prepare for the convex relaxation, for each $\mR_i=\Mat{\mR_i^{(1)}&\mR_i^{(2)}&\mR_i^{(3)}}\in\SO{3}$, we define the variable
\begin{equation}\label{eq:Ystructure}
  \begin{aligned}
    \mY_{i} = \Mat{\mR_{i}^{(1)}\\\mR_{i}^{(2)}\\1}\Mat{\mR_{i}^{(1)}\\\mR_{i}^{(2)}\\1}^\trans\in\IR^{7\times7}.
  \end{aligned}
\end{equation}

There exists a linear transformation from $\mY_i$ to $\mR_i$. The first two columns, $\mR_i^{(1)}$ and $\mR_i^{(2)}$, can be recovered from the last column of $\mY_i$ and the third column $\mR_i^{(3)}=\mR_i^{(1)}\times \mR_i^{(2)}$ is a linear function of $\mY_i$. We denote $g(\cdot)$ as the transformation from $\mY=\Mat{\mY_1&\mY_2&\dots &\mY_i &\dots &\mY_{n_r}}$ to $\mR$ such that $g(\mY)=\vec(\mR)$.

We define the following relaxation of $\cU$.
\begin{definition}
  The set $\bar{\cU}$ is defined as all $\bar{\vu}=g(\mY) \in\IR^{9n_r}$ such that
  \begin{subequations}\label{eq:defsdprelaxset}
    \begin{align}
      &\mA_{\textrm{axis}}\bar{\vu} = \vb_{\textrm{axis}},\label{eq:ubar axis}\\
      &\mA_{\textrm{angle}}\bar{\vu} \leq \vb_{\textrm{angle}},\label{eq:ubar angle}\\
      &\mA_{\textrm{structure}}\vec(\mY)=\vb_{\textrm{structure}}\label{eq:ubar structure}\\
      &\mY_{i}\succeq0,\forall i\in\cV_r\label{eq:ubar psd}
    \end{align}
  \end{subequations}
  where \eqref{eq:ubar structure} is a constraint that imposes the following structure on $\mY$ (resulted by combining \eqref{eq:ik manifold}) and \eqref{eq:Ystructure})
  \begin{enumerate}
  \item $\trace(\mY_i(1:3,1:3))=\trace(\mY_i(4:6,4:6))=1$;
  \item $\trace(\mY_i(1:3,4:6))=0$;
  \item $\mY_i(7,7)=1$.
  \end{enumerate}
\end{definition}
The kinematic constraints $\cU$ can be exactly captured by the relaxed set $\bar{\cU}$ intersected with a rank-1 constraint, as detailed in the fowllowing theorem (see \cite{wu2023cdc} for a proof).
\begin{theorem}\label{thm:intersect u_bar rank1}
  The set $\cU$ is a subset of $\bar{\cU}$, and is equal to the intersection of $\bar{\cU}$ and $\cR_1$, i.e., $\cU=\bar{\cU}\intersect \cR_1$, where $\cR_1$ is the set of $\vu=g(\mY)\in\IR^{9 n_r}$ such that each $\mY_{i}\in\real{7\times7}$ of $\mY$ is rank-1.
\end{theorem}

\subsection{Prismatic Joints}
The vision-based constraints use a formulation similar to that of the kinematic constraints for prismatic joints. A way to formulate such constraints is introduced in \cite{wu2024tac_archive}. This subsection gives a brief review.

Let $\ave_p\subset \ave$ represents the set of prismatic joints and $\cV_p$ the parents of prismatic joints. The prismatic joint can be defined as the following.
\begin{definition}\label{def:prismatic}
    A prismatic joint $(i,j)\in\ave_p$ is defined by the constraints:
    \begin{equation}\label{eq:prismatic_trans}
    \begin{aligned}
        &\mR_i=\mR_j, \\
        &\mR_i,\mR_j\in \SO{3}\\
        &\mT_j = \mT_i+(\tau_l+\tau_i(\tau_u-\tau_l))\mR_i^{(3)},\\
        &\tau_i\in [0,1],
    \end{aligned}
    \end{equation}
where $\tau_l,\tau_u$ are the lower- and upper-bound of the extension of the joint $\tau_i$. The prismatic joint is assumed to be aligned along the $z$-axis in this definition.
\end{definition}

To write convex constraints for \eqref{eq:prismatic_trans}, we define the variable $\mY_\tau=\Mat{\mY_{\tau 1}&\mY_{\tau 2} &\dots &\mY_{\tau i}&\dots &\mY_{\tau n_p}}$, and
\begin{equation}\label{eq:Y_itau definition}
    \begin{aligned}
       \mY_{\tau i} &= \Mat{\sqrt{\tau_i}\mR_{i}^{(3)}\\\sqrt{1-\tau_i}\mR_{i}^{(3)}\\\sqrt{\tau_i}\\\sqrt{1-\tau_i}}\Mat{\sqrt{\tau_i}\mR_{i}^{(3)}\\\sqrt{1-\tau_i}\mR_{i}^{(3)}\\\sqrt{\tau_i}\\\sqrt{1-\tau_i}}^\trans\in\IR^{8\times8}.
    \end{aligned}
\end{equation}
We define the constraint $\mY,\mY_\tau\in \bar{\cY}_\tau$ 
to restrict the following linear relations of $\mY_{\tau i}$ and $\mY_i$ entries: for $(i,j)\in\ave_p$,
\begin{enumerate}
    \item the trace of $\mY_{\tau i}$ equals 2;\label{itm:ytau1}
    \item $\trace(\mY_{\tau i}(1:3,1:3))=\mY_{i
    \tau}(7,7)$ and $\trace(\mY_{\tau i}(4:6,4:6))=\mY_{i
    \tau}(8,8)$;\label{itm:ytau2}
      \item $\mY_{\tau i}(4:6,7)=\mY_{\tau i}(1:3,8)$;\label{itm:ytau3}
      \item $\trace(\mY_{\tau i}(1:3,4:6))=\mY_{\tau i}(7,8)$;\label{itm:ytau4}
      \item $\mY_{\tau i}(7,7)\in [0,1]$;\label{itm:ytau5}
      \item $\mY_{\tau i}(7,8)\geq 0$;\label{itm:ytau6}
      \item $\mY_{\tau i}(1:3,7)+\mY_{\tau i}(4:6,8)=\Mat{\mY_i(2,6)-\mY_i(3,5)\\\mY_i(3,4)-\mY_i(1,6)\\\mY_i(1,5)-\mY_i(2,4)}$.\label{itm:ytau7}
\end{enumerate}

The above constraints restricts the structure of $\mY_{\tau i}$, its relation to $\mY_i$, and the bound on $\tau_i$. These constraints are utilized in the following theorem (see \cite{wu2024tac_archive} for a proof).
\begin{theorem}\label{thm:prismatic trans Y_tau}
    Equations \eqref{eq:prismatic_trans} hold if and only if
    \begin{equation}\label{eq:prismatic_trans_Y_tau}
        \mT_j = \mT_i+\tau_l\mR^{(3)}_i+(\tau_u-\tau_l)\mY_{\tau i}(1:3,7),
    \end{equation}
    $\mY_i$, $\mY_{\tau i}\in\IS_+$, $\mY,\mY_\tau\in \bar{\cY}_\tau$, $\mY_i$ satisfies \eqref{eq:ubar structure}, $\mR_i=\mR_j$, and $\rank(\mY_{i})=\rank(\mY_{\tau i}) = 1$.
\end{theorem}

\subsection{Inverse kinematics and  rank-constrained optimization}

Theorem \ref{thm:intersect u_bar rank1} and \ref{thm:prismatic trans Y_tau} enables us to write \eqref{eq:def ik feas set} and \eqref{eq:prismatic_trans} as a semidefinite constraint plus a rank-1 constraint on $\mY,\mY_\tau$. The IK problem can be formulated as the following
\begin{prob}[1][Rank constrained inverse kinematics]\label{prob:IK}
  \begin{subequations}\label{eq:ikrcproblem}
    \begin{align}
      &\min_{\mY,\mY_{\tau}} && f_{\textrm{ik}}(\mY,\mY_\tau)\\
      &\subjectto && g(\mY)\in\bar{\cU}, \label{eq:IKrc u}\\
      & &&\mY,\mY\in \bar{\cY}_\tau,\label{eq:IKrc yytau}\\
      & &&\mY_i=\mY_j, \forall(i,j)\in\ave_p,\label{eq:IKrc parallel}\\
      & &&\rank(\mY_{i}) = 1,\ i \in \cV_r, \label{eq:IKrc Y rank}\\
      & &&\rank(\mY_{\tau i}) = 1,\ i \in \cV_p. \label{eq:IKrc Ytau rank}
    \end{align}
  \end{subequations}
\end{prob}
where $f_{\textrm{ik}}$ is a quadratic function indicating the squared distance between poses of the end-effector and the target. The constraints \eqref{eq:IKrc u} and \eqref{eq:IKrc Y rank} ensure that $g(\mY)\in\cU$ using Theorem \ref{thm:intersect u_bar rank1}. The constraints \eqref{eq:IKrc parallel}, \eqref{eq:IKrc yytau}, and \eqref{eq:IKrc Ytau rank} together enforce the kinematic constraints for the prismatic joints defined in \eqref{eq:prismatic_trans} hold using Theorem \ref{thm:prismatic trans Y_tau}. The rank constraints \eqref{eq:IKrc Y rank} and \eqref{eq:IKrc Ytau rank} are nonconvex, but can be accounted for using a rank minimization algorithm. For example, \cite{wu2023cdc} provides a method and \cite{wu2024tac_archive} introduces an alternative approach.

\section{Visual Inverse Kinematics}
This section introduces the visibility constraints along with different vision-based costs.
\subsection{Perspective projection and virtual prismatic joints}\label{sec:perspective proj}
\begin{figure}[h]
  \centering
  \resizebox{0.42\textwidth}{!}{
  \tikzset{every picture/.style={line width=0.75pt}} 

\begin{tikzpicture}[x=0.75pt,y=0.75pt,yscale=-1,xscale=1]

\draw  [color={rgb, 255:red, 248; green, 231; blue, 28 }  ,draw opacity=1 ][fill={rgb, 255:red, 248; green, 231; blue, 28 }  ,fill opacity=1 ] (474.37,192.67) -- (213,278.27) -- (412.37,363.33) -- (455.53,355.33) -- (521,237.54) -- cycle ;
\draw    (212.96,278.27) -- (369.45,328) ;
\draw [color={rgb, 255:red, 248; green, 231; blue, 28 }  ,draw opacity=1 ][fill={rgb, 255:red, 248; green, 231; blue, 28 }  ,fill opacity=1 ]   (213.04,278.27) -- (474.37,192.67) ;
\draw [color={rgb, 255:red, 248; green, 231; blue, 28 }  ,draw opacity=1 ][fill={rgb, 255:red, 248; green, 231; blue, 28 }  ,fill opacity=1 ]   (474.37,192.67) .. controls (517.71,183.33) and (521.71,224) .. (521,237.54) ;
\draw [color={rgb, 255:red, 248; green, 231; blue, 28 }  ,draw opacity=1 ][fill={rgb, 255:red, 248; green, 231; blue, 28 }  ,fill opacity=1 ]   (521,237.54) .. controls (521.71,266.67) and (504.37,322.67) .. (455.53,355.33) ;
\draw [color={rgb, 255:red, 248; green, 231; blue, 28 }  ,draw opacity=1 ][fill={rgb, 255:red, 248; green, 231; blue, 28 }  ,fill opacity=1 ]   (455.53,355.33) .. controls (433.04,368.67) and (429.04,366.67) .. (412.37,363.33) ;
\draw [color={rgb, 255:red, 248; green, 231; blue, 28 }  ,draw opacity=1 ][fill={rgb, 255:red, 248; green, 231; blue, 28 }  ,fill opacity=1 ]   (213,278.27) -- (412.37,363.33) ;

\draw [line width=2.25]    (212.96,278.27) -- (455.49,278.27) ;
\draw  [fill={rgb, 255:red, 255; green, 255; blue, 255 }  ,fill opacity=0.59 ] (541.53,301.74) -- (369.45,408.93) -- (369.45,408.93) -- (369.45,254.8) -- (541.53,147.6) -- cycle ;
\draw    (212.96,278.27) -- (369.45,262.27) ;
\draw [line width=2.25]    (212.96,278.27) -- (212.96,170.33) ;
\draw [shift={(212.96,165.33)}, rotate = 90] [fill={rgb, 255:red, 0; green, 0; blue, 0 }  ][line width=0.08]  [draw opacity=0] (14.29,-6.86) -- (0,0) -- (14.29,6.86) -- cycle    ;
\draw [line width=2.25]    (212.96,278.27) -- (337.78,200.51) ;
\draw [shift={(342.02,197.87)}, rotate = 148.08] [fill={rgb, 255:red, 0; green, 0; blue, 0 }  ][line width=0.08]  [draw opacity=0] (14.29,-6.86) -- (0,0) -- (14.29,6.86) -- cycle    ;
\draw [line width=2.25]    (212.96,278.27) -- (126.92,331.86) ;
\draw [line width=2.25]    (455.49,278.27) -- (455.49,355.33) ;
\draw    (455.49,201.2) -- (455.49,278.27) ;
\draw [line width=2.25]    (455.49,278.27) -- (369.45,331.86) ;
\draw    (541.53,224.67) -- (455.49,278.27) ;
\draw [line width=2.25]    (455.49,278.27) -- (455.49,178.33) ;
\draw [shift={(455.49,173.33)}, rotate = 90] [fill={rgb, 255:red, 0; green, 0; blue, 0 }  ][line width=0.08]  [draw opacity=0] (14.29,-6.86) -- (0,0) -- (14.29,6.86) -- cycle    ;
\draw [line width=2.25]    (455.49,278.27) -- (558.79,213.91) ;
\draw [shift={(563.04,211.27)}, rotate = 148.08] [fill={rgb, 255:red, 0; green, 0; blue, 0 }  ][line width=0.08]  [draw opacity=0] (14.29,-6.86) -- (0,0) -- (14.29,6.86) -- cycle    ;
\draw  [draw opacity=0][fill={rgb, 255:red, 0; green, 0; blue, 0 }  ,fill opacity=1 ] (627.83,238.17) .. controls (627.83,236.23) and (629.4,234.67) .. (631.33,234.67) .. controls (633.27,234.67) and (634.83,236.23) .. (634.83,238.17) .. controls (634.83,240.1) and (633.27,241.67) .. (631.33,241.67) .. controls (629.4,241.67) and (627.83,240.1) .. (627.83,238.17) -- cycle ;
\draw  [draw opacity=0][fill={rgb, 255:red, 0; green, 0; blue, 0 }  ,fill opacity=1 ] (428.67,256.83) .. controls (428.67,254.9) and (430.23,253.33) .. (432.17,253.33) .. controls (434.1,253.33) and (435.67,254.9) .. (435.67,256.83) .. controls (435.67,258.77) and (434.1,260.33) .. (432.17,260.33) .. controls (430.23,260.33) and (428.67,258.77) .. (428.67,256.83) -- cycle ;
\draw    (432.17,256.83) -- (631.33,238.17) ;
\draw   (455.49,355.33) .. controls (419.32,377.83) and (390.01,361.55) .. (390.01,318.99) .. controls (390.01,276.43) and (419.32,223.69) .. (455.49,201.2) .. controls (491.65,178.71) and (520.96,194.98) .. (520.96,237.54) .. controls (520.96,280.1) and (491.65,332.84) .. (455.49,355.33) -- cycle ;
\draw [line width=2.25]    (455.49,278.27) -- (596,278.27) ;
\draw [shift={(601,278.27)}, rotate = 180] [fill={rgb, 255:red, 0; green, 0; blue, 0 }  ][line width=0.08]  [draw opacity=0] (14.29,-6.86) -- (0,0) -- (14.29,6.86) -- cycle    ;
\draw  [dash pattern={on 4.5pt off 4.5pt}]  (369.45,262.27) -- (432.17,256.83) ;
\draw    (277.49,278.27) .. controls (283.67,285.33) and (282.33,295.33) .. (277.49,298.67) ;
\draw  [dash pattern={on 4.5pt off 4.5pt}]  (369.45,328) -- (455.49,355.33) ;

\draw (288,282.67) node [anchor=north west][inner sep=0.75pt]    {$\alpha $};
\draw (442.17,363.53) node [anchor=north west][inner sep=0.75pt]  [rotate=-329.47] [align=left] {image plane};
\draw (191.33,295) node [anchor=north west][inner sep=0.75pt]   [align=left] {camera \\center};
\draw (335.33,162.4) node [anchor=north west][inner sep=0.75pt]    {$X$};
\draw (216,145.07) node [anchor=north west][inner sep=0.75pt]    {$Y$};
\draw (447.33,145.4) node [anchor=north west][inner sep=0.75pt]    {$y$};
\draw (568,196.73) node [anchor=north west][inner sep=0.75pt]    {$x$};
\draw (594.67,288.4) node [anchor=north west][inner sep=0.75pt]    {$Z$};
\draw (634.33,218.57) node [anchor=north west][inner sep=0.75pt]    {$P$};
\draw (429.33,228.07) node [anchor=north west][inner sep=0.75pt]    {$p$};

\end{tikzpicture}
  }
  \caption{The field of view is represented as a right circular cone whose axis is aligned with the camera Z-axis.}
  \label{fig:pinhole}
\end{figure}
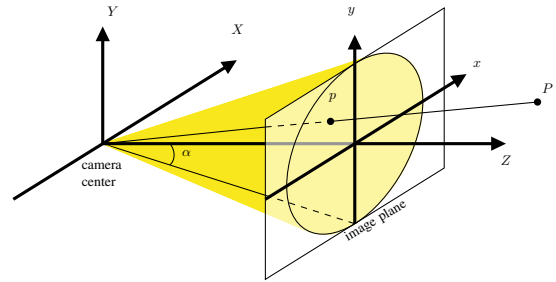

In this work we use the pinhole camera model \cite{hartley2003multiple}. We denote $\mR_{\textrm{c}}$ as the rotation matrix of the reference frame whose axes are aligned with the camera axes $X,Y$ and $Z$. As visualized in Fig. \ref{fig:pinhole}, a point $\mP\in\IR^3$ is projected to a point $\vp\in\IR^2$ on the image plane by the transformation $\pi:\IR^3\mapsto\IR^2$,
\begin{equation}
    \vp\sim\pi(\mP):=\mC\Mat{\mP\\ 1},
\end{equation}
where $\mC$ is the camera matrix. In order to capture the point $\mP$ in the field of view, $\vp$ must be within the rectangular limit of the digital image. In this work, we reduce the field of view by limiting $\vp$ in a circle centered at the principal point with radius $h/2$, where $h$ is the height of the image. In a 3-D world, the field of view then becomes a circular cone (assuming the height is infinitely large) whose apex is located at camera center, and the axis is aligned with the camera $Z$-axis. We define $\alpha$ as the aperture (or half-angle) of the cone and $\alpha = r_\alpha\arctan(\frac{h}{2f_c})$, where $f_c$ is the focal length and $r_\alpha\in(0,1]$ is a factor to control the tightness of the FoV constraint.

\begin{figure}[h]
      \centering
      \resizebox{0.42\textwidth}{!}{
  \input{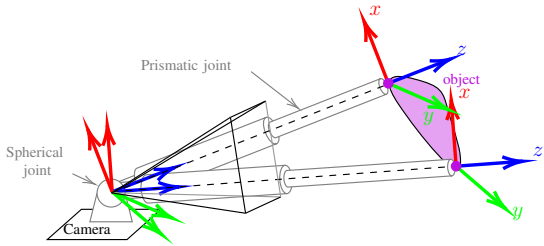}\label{fig:ref_frames_obstacle}
  }
  \caption{The object is connected with the camera center through a virtual chain consisted of prismatic joints and spherical joints. Each prismatic joint is parameterized with two reference frames sharing the same $z$-axis.}
  \label{fig:fov_model}
\end{figure}

The visibility constraint ensures that the target stays in the field-of-view of the camera. To include this constraint in our formulation, we first model the robot and the filmed object using reference frames and then develop constraints by placing \emph{virtual} joints between them.
To begin with, we assume that the 3-D position of the object is available and that the object is characterized with a set of points $\cV_o$, the position of which is refered as $\{\mT_i\}_{i\in\cV_o}$ (for simplicity, we do not consider self-occlusion for the object). For each point $i\in\cV_o$, we connect the point to the camera center (denoted as the index $\textrm{c}$) through a series of virtual joints: starting from the point, we put a prismatic joint followed by a spherical joint located at the camera center, as shown in Fig.\ref{fig:fov_model}. Then we place two reference frames, attached to each link of the prismatic joint. 

\begin{remark}
Ideally, the field of view should have the shape of a rectangular pyramid since images are rectangular. However, the orientation of this shape is a function of the pose of the camera, which results in more involved FoV constraints. It should be possible to formulate also the constraint in our IK framework by introducing two additional coincident revolute joints at the camera center in addition to the prismatic joint. However, we decide to keep the scope of the paper focused on the simpler VIK problem with a conic FoV.
\end{remark}

\subsection{Visibility constraint}
\begin{figure}[h]
  \centering
  \resizebox{0.32\textwidth}{!}{
  \tikzset{every picture/.style={line width=0.75pt}} 

\begin{tikzpicture}[x=0.75pt,y=0.75pt,yscale=-1,xscale=1]

\draw  [draw opacity=0][fill={rgb, 255:red, 248; green, 231; blue, 28 }  ,fill opacity=0.6 ] (213,278.27) -- (524.88,127.01) -- (524.88,429.52) -- cycle ;
\draw  [draw opacity=0][fill={rgb, 255:red, 248; green, 231; blue, 28 }  ,fill opacity=0.6 ] (524.88,127.01) .. controls (535.11,127.59) and (543.39,195.08) .. (543.39,278.26) .. controls (543.39,361.8) and (535.04,429.52) .. (524.74,429.52) -- (524.74,278.26) -- cycle ;
\draw [line width=2.25]    (213,278.27) -- (213,173) ;
\draw [shift={(213,168)}, rotate = 90] [fill={rgb, 255:red, 0; green, 0; blue, 0 }  ][line width=0.08]  [draw opacity=0] (14.29,-6.86) -- (0,0) -- (14.29,6.86) -- cycle    ;
\draw [line width=2.25]    (213,278.27) -- (563,278.27) ;
\draw [shift={(568,278.27)}, rotate = 180] [fill={rgb, 255:red, 0; green, 0; blue, 0 }  ][line width=0.08]  [draw opacity=0] (14.29,-6.86) -- (0,0) -- (14.29,6.86) -- cycle    ;
\draw    (213,278.27) -- (524.88,127.01) ;
\draw    (213,278.27) -- (524.88,429.52) ;
\draw  [draw opacity=0][fill={rgb, 255:red, 144; green, 19; blue, 254 }  ,fill opacity=0.36 ] (310.36,278.27) .. controls (310.36,234.01) and (346.24,198.13) .. (390.5,198.13) .. controls (434.76,198.13) and (470.64,234.01) .. (470.64,278.27) .. controls (470.64,322.53) and (434.76,358.4) .. (390.5,358.4) .. controls (346.24,358.4) and (310.36,322.53) .. (310.36,278.27) -- cycle ;
\draw [color={rgb, 255:red, 0; green, 0; blue, 255 }  ,draw opacity=1 ][line width=3]    (213,278.27) -- (314.79,239.27) -- (374.09,216.56) ;
\draw [shift={(378.75,214.77)}, rotate = 159.04] [color={rgb, 255:red, 0; green, 0; blue, 255 }  ,draw opacity=1 ][line width=3]    (20.77,-6.25) .. controls (13.2,-2.65) and (6.28,-0.57) .. (0,0) .. controls (6.28,0.57) and (13.2,2.66) .. (20.77,6.25)   ;
\draw [color={rgb, 255:red, 65; green, 117; blue, 5 }  ,draw opacity=1 ][line width=3]    (213,278.27) -- (322,278.27) -- (385.5,278.27) ;
\draw [shift={(390.5,278.27)}, rotate = 180] [color={rgb, 255:red, 65; green, 117; blue, 5 }  ,draw opacity=1 ][line width=3]    (20.77,-6.25) .. controls (13.2,-2.65) and (6.28,-0.57) .. (0,0) .. controls (6.28,0.57) and (13.2,2.66) .. (20.77,6.25)   ;
\draw [color={rgb, 255:red, 144; green, 19; blue, 254 }  ,draw opacity=1 ][line width=3]    (390.5,278.27) -- (379.66,219.69) ;
\draw [shift={(378.75,214.77)}, rotate = 79.52] [color={rgb, 255:red, 144; green, 19; blue, 254 }  ,draw opacity=1 ][line width=3]    (20.77,-6.25) .. controls (13.2,-2.65) and (6.28,-0.57) .. (0,0) .. controls (6.28,0.57) and (13.2,2.66) .. (20.77,6.25)   ;
\draw  [draw opacity=0][dash pattern={on 4.5pt off 4.5pt}] (524.88,429.52) .. controls (524.88,429.52) and (524.88,429.52) .. (524.88,429.52) .. controls (514.93,429.52) and (506.88,361.8) .. (506.88,278.27) .. controls (506.88,194.73) and (514.93,127.01) .. (524.88,127.01) -- (524.88,278.27) -- cycle ; \draw  [dash pattern={on 4.5pt off 4.5pt}] (524.88,429.52) .. controls (524.88,429.52) and (524.88,429.52) .. (524.88,429.52) .. controls (514.93,429.52) and (506.88,361.8) .. (506.88,278.27) .. controls (506.88,194.73) and (514.93,127.01) .. (524.88,127.01) ;  
\draw  [draw opacity=0] (524.74,127.01) .. controls (524.74,127.01) and (524.74,127.01) .. (524.74,127.01) .. controls (534.68,127.01) and (542.74,194.73) .. (542.74,278.26) .. controls (542.74,361.8) and (534.68,429.52) .. (524.74,429.52) -- (524.74,278.26) -- cycle ; \draw   (524.74,127.01) .. controls (524.74,127.01) and (524.74,127.01) .. (524.74,127.01) .. controls (534.68,127.01) and (542.74,194.73) .. (542.74,278.26) .. controls (542.74,361.8) and (534.68,429.52) .. (524.74,429.52) ;  
\draw  [draw opacity=0] (285.69,243.64) .. controls (290.7,254.13) and (293.5,265.87) .. (293.5,278.27) .. controls (293.5,278.51) and (293.5,278.75) .. (293.5,279) -- (213,278.27) -- cycle ; \draw   (285.69,243.64) .. controls (290.7,254.13) and (293.5,265.87) .. (293.5,278.27) .. controls (293.5,278.51) and (293.5,278.75) .. (293.5,279) ;  

\draw (563,282.67) node [anchor=north west][inner sep=0.75pt]    {$Z$};
\draw (337,185.91) node [anchor=north west][inner sep=0.75pt]  [color={rgb, 255:red, 0; green, 0; blue, 255 }  ,opacity=1 ]  {$\mathbf{R}_{i}^{( 3)}$};
\draw (294.67,248.07) node [anchor=north west][inner sep=0.75pt]    {$\alpha $};
\draw (376,288.41) node [anchor=north west][inner sep=0.75pt]  [color={rgb, 255:red, 0; green, 0; blue, 255 }  ,opacity=1 ]  {$\mathbf{\textcolor[rgb]{0.25,0.46,0.02}{R}}\textcolor[rgb]{0.25,0.46,0.02}{_{c}^{( 3)}}$};

\end{tikzpicture}
  }
  \caption{The visibility constraint formulated as a ball bound on the difference $\mR_i^{(3)}-\mR^{(3)}_{\textrm{c}}$.}
  \label{fig:cone}
\end{figure}
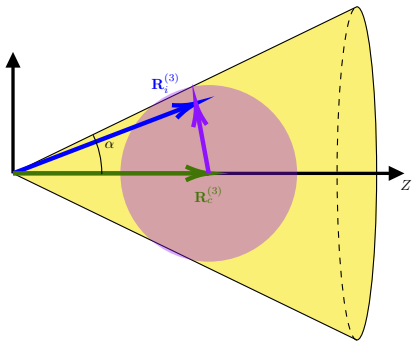

From definition \ref{def:prismatic}, the $z$-axes of the two frames attached to the links of the prismatic joint are aligned and pass through the camera center $\textrm{c}$ and the point $i\in\cV_o$. As a result, the field of view constraint is the same as restricting this $z$-axis in the cone, which is equivalent to enforcing the ball bound:
\begin{equation}\label{eq:ball bound}
    \begin{aligned}
        \mR^{(3)}_i-\mR_{\textrm{c}}^{(3)}\in \cS(\sqrt{2-2\cos(\alpha)}),
    \end{aligned}
\end{equation}
where $\mR_i^{(3)}$ and $\mR^{(3)}_{\textrm{c}}$ are the third columns coincide with the $z$-axis of the two frames, $\cS(r)$ is a ball centered at the origin with radius $r$. When $\mR^{(3)}_i$ is on the cone surrounding $\mR_{\textrm{c}}^{(3)}$, the two unit vectors form an isosceles triangle with the apex angle equals $\alpha$ and legs equals $1$. The length of the triangle base can be found as $2\sin(\alpha/2)=\sqrt{2-2\cos(\alpha)}$. It is clear that $\mR^{(3)}_i$ remains in the cone when $\mR^{(3)}_i-\mR_{\textrm{c}}^{(3)}\in \cS(\sqrt{2-2\cos(\alpha)})$. 
This constraint is equivalent to the joint angle constraint \eqref{eq:ik angle} in \cite{wu2023cdc}, where we restrict the difference of the first columns of the two neighboring reference frames in a ball. A visualization of \eqref{eq:ball bound} is in Fig. \ref{fig:cone}. 

We formulate the VIK problem similarly as we do for the IK problem of the robot with additional virtual joints, which is
\begin{prob}[2][Visual inverse kinematics]\label{prob:VIK}
  \begin{subequations}\label{eq:vikproblem}
    \begin{align}
      &\min_{\mY,\mY_{\tau}} && f_{\textrm{vik}}(\mY,\mY_\tau)\label{eq:vik cost}\\
      &\subjectto &&\mY,\mY_\tau \text{ satisfy \eqref{eq:IKrc u}-\eqref{eq:IKrc Ytau rank}},\label{eq:VIK constraint}\\
      & &&\mA_{\textrm{cone}}\mY\leq \vb_{\textrm{cone}}.\label{eq:vik cone}
    \end{align}
  \end{subequations}
\end{prob}
In Problem \ref{prob:VIK}, the objective function \eqref{eq:vik cost} is a vision-based cost, which we discuss in the next section. The constraint \eqref{eq:VIK constraint} contains all the constraints in Problem \ref{prob:IK} and captures the kinematics constraints of the robot while \eqref{eq:vik cone} 
is a relaxation of the visibility constraint \eqref{eq:ball bound}, where the ball restricting $\mR_i^{(3)}-\mR_{\textrm{c}}^{(3)}$ is approximated with a polyhedron represented by a linear matrix inequality. Eventually, because $\mR_i^{(3)}-\mR_{\textrm{c}}^{(3)}$ is a function of $\mY$, \eqref{eq:vik cone} is obtained as a linear constraint on $\mY$.

\subsection{Vision-based costs}\label{sec:costs}
The cost \eqref{eq:vik cost} in Problem \ref{prob:VIK} can have different forms to fulfill different visual tasks. 
In this section, we introduce three costs that encode different objectives of the visual inverse kinematics problem. To perform flexible tasks, these costs can be used jointly by combining them linearly.
\subsubsection{Levelness}
When generating vision data, it is by convention that the image is level with the ground. Therefore it is important to find robot joint configurations that can make such images. To encode this objective in the cost we assume that, as customary, the $y$-axis of the camera image plane represents the upward direction of the image. We then propose the objective function
\begin{equation}\label{:level}
    \begin{aligned}
        f(\mR) = \norm{\mR^{(2)}_{\textrm{c}}-\Mat{0 &0 &1}^\trans}_2^2,
    \end{aligned}
\end{equation}
which evaluate the difference between the second column of $\mR_{\textrm{c}}$ and the world z-axis frame. When minimized, the cost indicates that the camera is close to an upright condition.

\subsubsection{Centering}
The visibility constraints keep the object in the image, but do not specify \emph{where} in the image. In some scenarios we would like to keep the object around the center of the image, motivating the following cost function.

\begin{equation}\label{eq:concentrate}
    \begin{aligned}
        f(\mR) = \sum_{i\in\cV_o}\norm{\mR^{(3)}_i-\mR^{(3)}_{\textrm{c}}}_2^2
    \end{aligned}
\end{equation}
Function \ref{eq:concentrate} evaluates the sum of the squared distance between $\mR^{(3)}_i$ and $\mR_{\textrm{c}}^{(3)}$. Minimizing \eqref{eq:concentrate} means that the vector in the l.h.s of \eqref{eq:ball bound} is not only bounded in the ball, but also minimized for all the points $\cV_o$. We can also see that minimizing \eqref{eq:concentrate} is equivalent to maximizing $\sum_{i\in\cV_o}\mR^{(3)\trans}_{\textrm{c}}\mR_i^{(3)}=\sum_{i\in\cV_o}\cos(\theta_{i\textrm{c}})$, where $\theta_{i\textrm{c}}$ is the angle between the two vectors. Therefore by minimizing \eqref{eq:concentrate}, the vectors $\{\mR_i^{(3)}\}_{i\in\cV_o}$ are averaged such that the vector $\mR_{\textrm{c}}^{(3)}$ is as close as possible to the mean on the unit sphere (interested readers are refered to \cite{hartley2013rotation} for averaging rotations on manifold). 

Minimizing \eqref{eq:concentrate} is the same as maximizing $\sum_{i\in\cV_o}\cos(\theta_{i\textrm{c}})$, thus pushing the camera away from the points. We provide below another centering cost that, when minimized, yields camera poses that are close to the objects. 

\begin{equation}\label{eq:concentrate_close}
    \begin{aligned}
        f(\mR,\mT) = \sum_{i\in\cV_o}\norm{\mT_i-\mT_{\textrm{c}}-\mR^{(3)}_{\textrm{c}}}_2^2
    \end{aligned}
\end{equation}

Minimizing \eqref{eq:concentrate_close} as the cost in Problem \ref{prob:VIK} finds a configuration such that the camera is as close as possible to the object points while centering them in the image. 

\subsubsection{Minimizing the reprojection error}
The reprojection error is the distance between the projection of a point $\mP$ onto the image plane, $\vp$, and a corresponding measurement $\hat{\vp}$. Suppose we are given a preexisting image of the object, and would like to find the configuration such that the reprojection error of the camera is minimized (e.g., as part of a standardization of an industrial inspection procedure). We denote the coordinates of the projections of the object points as $\{\hat{\vp}_i\}_{i\in\cV_o}$, and their corresponding 3-D vectors as $\tilde{\vp}_i=\Mat{\hat{\vp}_i\\f_c}$. The \emph{spherical} reprojection error is given by

\begin{equation}\label{eq:reproject error}
    \begin{aligned}
        f(\mR) = \sum_{i\in\cV_o}\norm{\mR^{(3)}_i-\frac{\tilde{\vp}_i}{\norm{\tilde{\vp}_i}}}_2^2,
    \end{aligned}
\end{equation}
where $\mR_i^{(3)}$ is a unit vector pointing toward the point $i$ and $\frac{\tilde{\vp}_i}{\norm{\tilde{\vp}_i}}$ is the unit vector that passes through the corresponding point on the image. Minimizing \eqref{eq:reproject error} finds a solution such that the camera can take a picture that matches the given image.

\subsection{Optimization}
This subsection introduces a strategy to solve the visual inverse kinematics problem. At first, a convex relaxation of the problem is solved. After that, the solution is moved iteratively to the set of rank-1 matrices through a rank minimization algorithm based on the work in \cite{wu2024tac_archive}.

\subsubsection{Convex relaxation}
Without the visibility constraint \eqref{eq:vik cone}, Problem \ref{prob:VIK} is the same as an inverse kinematics problem. Therefore we can perform the same convex relaxation as we do for robot inverse kinematics problem, which leads to the following formulation.
\begin{prob}[3][Relaxed Visual inverse kinematics]\label{prob:VIKrelax}
  \begin{subequations}\label{eq:vikrcproblem}
    \begin{align}
      &\min_{\mY,\mY_{\tau}} && f(\mY,\mY_\tau)\label{eq:VIKrelax obj}\\
      &\subjectto && g(\mY)\in\bar{\cU}, \label{eq:VIKrelax u}\\
      & &&\mY,\mY_\tau\in \bar{\cY}_\tau.\label{eq:VIKrelax yytau}\\
        & &&\mY_i=\mY_j, \forall(i,j)\in\ave_p\label{eq:VIKrelax parallel}\\
      & &&\mA_{\textrm{cone}}\mY\leq \vb_{\textrm{cone}}\label{eq:VIKrelax cone}
    \end{align}
  \end{subequations}
\end{prob}
The objective function \eqref{eq:VIKrelax obj} can be any of the costs in Section \ref{sec:costs} passed on to $\mY,\mY_\tau$.
Problem \ref{prob:VIKrelax} is a relaxation of Problem \ref{prob:VIK} because they are the same except that the rank constraints \eqref{eq:IKrc Y rank} and \eqref{eq:IKrc Ytau rank}, are omitted. We can apply off-the-shelf solver to Problem \ref{prob:VIKrelax} but the solution for $\mY_i$ and $\mY_{\tau i}$ is not, in general, rank-1. We briefly discuss below a way to project such solutions to the set of rank-1 matrices. 

\subsubsection{Rank minimization}
Two rank minimization algorithms are provided in \cite{wu2023cdc} and \cite{wu2024tac_archive}. The method in \cite{wu2023cdc} is based on the assumption that the minimal cost is zero and there exists an optimizer of this zero cost in the feasible set. The method in \cite{wu2024tac_archive} dose not require this assumption and can be applied to problems with any minimal cost. We apply the latter approach in this paper because the minimal values of the vision-based costs introduced in \ref{sec:costs} are generally non-zero.

The key idea in this rank minimization method is to maximize the largest eigenvalue of each $\mY_i$ and $\mY_{\tau i}$, and, because $\trace(\mY_i)$ and $\trace(\mY_{\tau i})$ are constant, the rest of the eigenvalues decrease accordingly and eventually render rank-1 matrices with only one non-zero eigenvalue. The method iteratively solves the following problem for an update $\{\mU^k,\mU^{k}_{\tau}\}$ of the variables $\{\mY^k,\mY^k_\tau\}$ at the $k$-th iteration.
\begin{prob}[4][Update problem]\label{prob:VIK update}
  \begin{subequations}
    \begin{align}
      &\min_{\mU^k,\mU^{k}_{\tau}} f(\mY^{k-1}+\mU^{k},\mY^{k-1}_{\tau}+\mU^{k}_{\tau})\\
          &\subjectto \notag\\
          &\sum_{i\in\cV_r}\vec(\mU^k_{i})^\trans\nabla\lambda_{1}(\mY^{k-1}_i) {\geq} \sum_{i\in\cV_r}(c{-}1)(\lambda_{1}(\mY^{k-1}_i){-}3) \label{eq:lin approx c Y}\\
          &\sum_{i\in\cV_p}\vec(\mU^{k-1}_{\tau i})^\trans\nabla\lambda_{1}(\mY^{k-1}_{\tau i}) {\geq} \sum_{i\in\cV_p}(c{-}1)(\lambda_{1}(\mY^k_{\tau i}){-}2)\label{eq:lin approx c Ytau}\\
          &\mY^{k-1}+\mU^k, \mY^{k-1}_{\tau}+\mU\tau^{k}\text{ satisfy \eqref{eq:VIKrelax u} - \eqref{eq:VIKrelax cone}.}\label{eq:IK4 Y}
    \end{align}
  \end{subequations}
\end{prob}
In Problem \ref{prob:VIK update}, $\lambda_{1}(\cdot)$ is the largest eigenvalue as a function of a matrix and $c\in[0,1]$ is a constant. The complete algorithm for solving the visual inverse kinemtics problem is presented in Algorithm \ref{alg:vik}. This algorithm uses an adaptive approach in steps \ref{step:startwhile2}-\ref{step:endwhile2} to find a constant $c$ that yields feasible solutions of Problem \ref{prob:VIK update}. By doing this, the success rate of the algorithm can be boosted (see \cite{wu2024tac_archive} for a compare).

\begin{algorithm}[h]
  \caption{Visual Inverse Kinematics Solver}\label{alg:vik}
  \hspace*{\algorithmicindent} \textbf{Input} $\{\mT_{i}\}_{i\in\cV_o}$, $\mu,\epsilon_1,\epsilon_2,k_{max},p_{max},c_0,a$\\
  \hspace*{\algorithmicindent} \textbf{Output} $\{\mR^*_i,\mT^*_i\}_{i\in\cV_r}$
  \begin{algorithmic}[1]
    \State Solve Problem \ref{prob:VIKrelax} to get an initial solution $\mY^0,\mY_{\tau}^0$ and set $k=1$, $p=1$.\label{step:init}
    \While{($\exists \lambda_{1}(\mY_i)\leq3-\epsilon_1$ $||$ $\exists\lambda_{1}(\mY_{\tau i})\leq2-\epsilon_1)$ $\&$ $\norm{\mU^k}_{F}\geq \epsilon_2$ $\&$ $k\leq k_{max}$}\label{step:startwhile}
    \State \parbox[t]{200pt}{For each $\mY^{k-1}_{i}$ and $\mY^{k-1}_{\tau i}$, compute the largest eigenvalues $\lambda_{1}(\mY_i)$ and $\lambda_{1}(\mY_{\tau i})$, respectively.}
    \While{$p\leq p_{max}$ $\&$ Problem \ref{prob:VIK update} is infeasible}\label{step:startwhile2}
    \State $c=1-(1-c_0)^{(a(p-1)+1)}$, $p\leftarrow p+1$.\label{step:c_update}
    \State \parbox[t]{200pt}{Solve Problem \ref{prob:VIK update} to get $\mU^{k}$ and $\mU^{k}_{\tau}$.}\label{step:update}
    \EndWhile\label{step:endwhile2}
    \State \parbox[t]{200pt}{Update $\mY^k_{i} = \mY^{k-1}_{i}+\mU^k_{i}$ for all $i\in\cV_r$ and update $\mY^k_{\tau i} = \mY^{k-1}_{\tau i}+\mU^k_{\tau i}$ for all $i\in\cV_p$ and set $k = k+1$, $p=1$.}
    \EndWhile\label{step:endwhile}
    \State Recover the rotations $\{\mR_{i}\}$ from $g(\mY^{k-1})$.\label{step:getrotations}
    \State Recover the translations $\{\mT_{i}\}$ using \eqref{eq:translation relation}.
  \end{algorithmic}
\end{algorithm}
Each iteration of Algorithm \ref{alg:vik} minimizes an upper bound on a concave cost, so it guarantees local convergence. See \cite{wu2024tac_archive} for a detailed proof of convergence.

\section{Simulation Results}
To validate the proposed method, simulations are performed on a 7-degrees-of-freedom Sawyer manipulator, which is mounted with a camera on its hand and tasked to take a photo of some quadrotors. We assume that the positions of the quadrotors are known and each quadrotor is simplified as a point. We connect each point with the camera center through the virtual joints mentioned in Section \ref{sec:perspective proj}. The visibility constraint requires the camera to capture the quadrotors in the field of view, i.e., to restrict the points in the viewing frustum. 

\begin{figure}[h]
    \centering
    \subfloat[Level]{
    \begin{minipage}[c][1\width]{
        0.24\textwidth}
      \centering
      \includegraphics[width=1\textwidth]{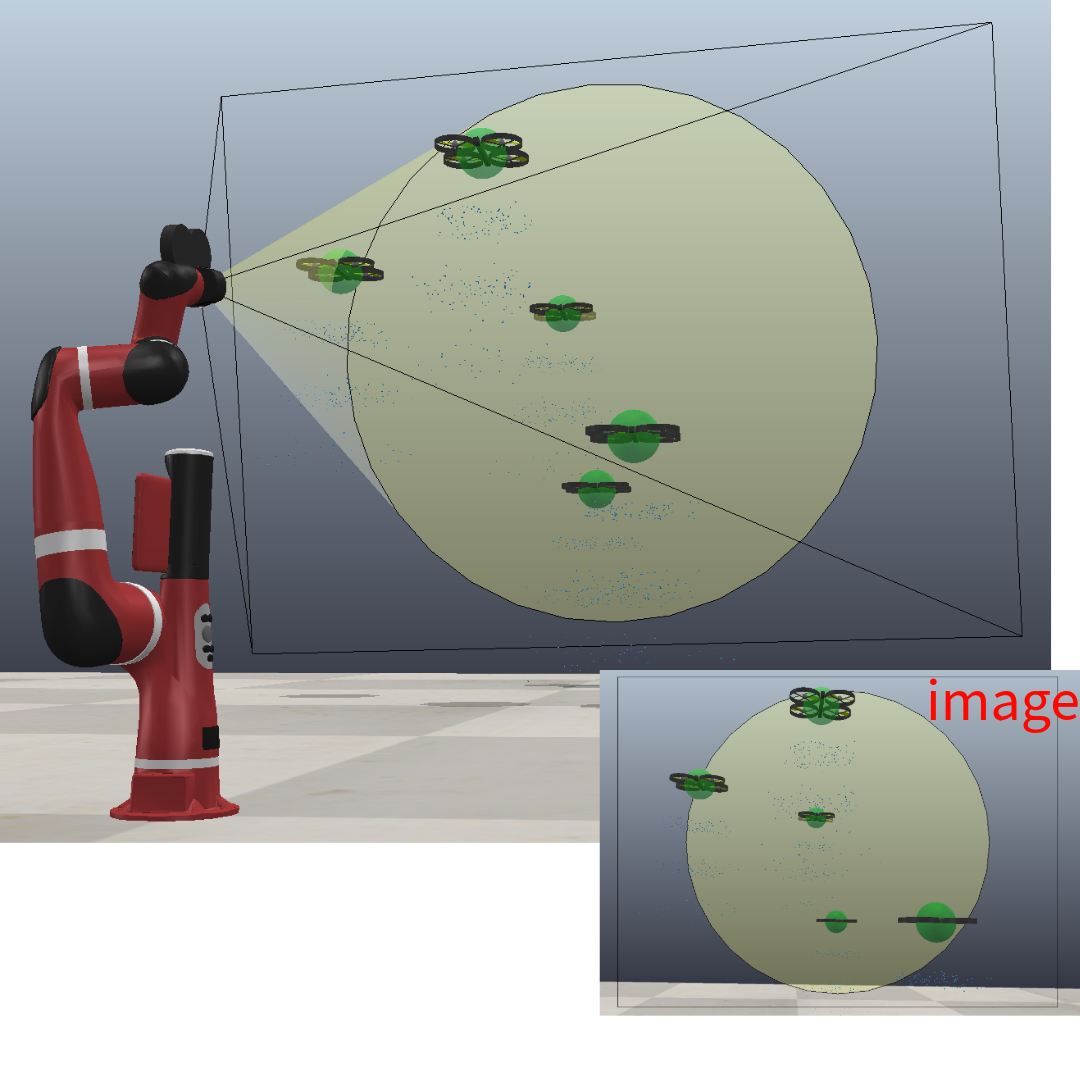}\label{fig:quad1}
    \end{minipage}}
    \subfloat[Level and center]{
    \begin{minipage}[c][1\width]{
        0.24\textwidth}
      \centering
      \includegraphics[width=1\textwidth]{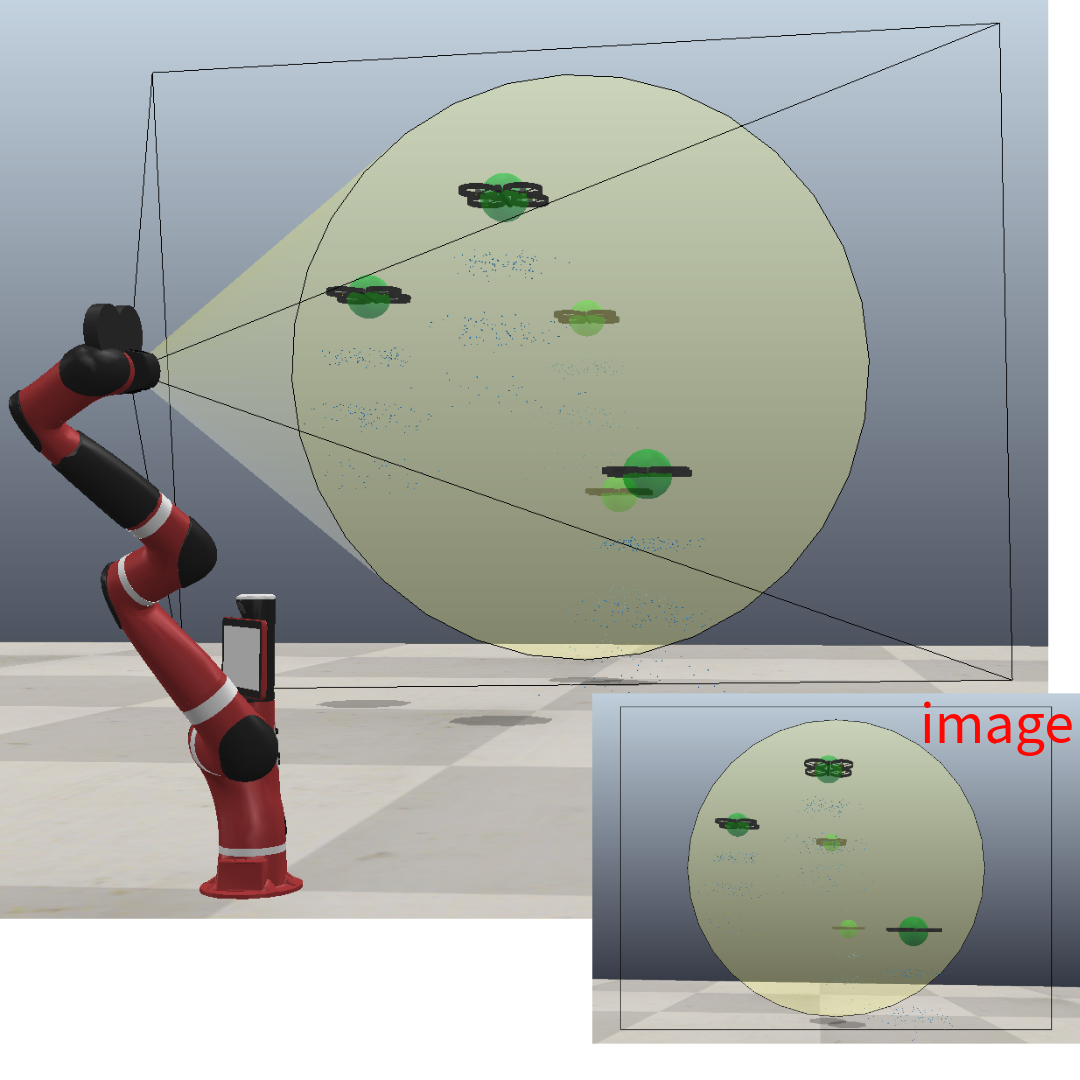}\label{fig:quad2}
    \end{minipage}}
    \caption{Our solver finds a posture of Sawyer subject to the visibility constraint of capturing the quadrotors in the camera field of view while targeting different vision-based objectives.}
    \label{fig:sawyer_quads}
\end{figure}

\begin{figure}[h]
    \centering
    \includegraphics[width=0.75\linewidth]{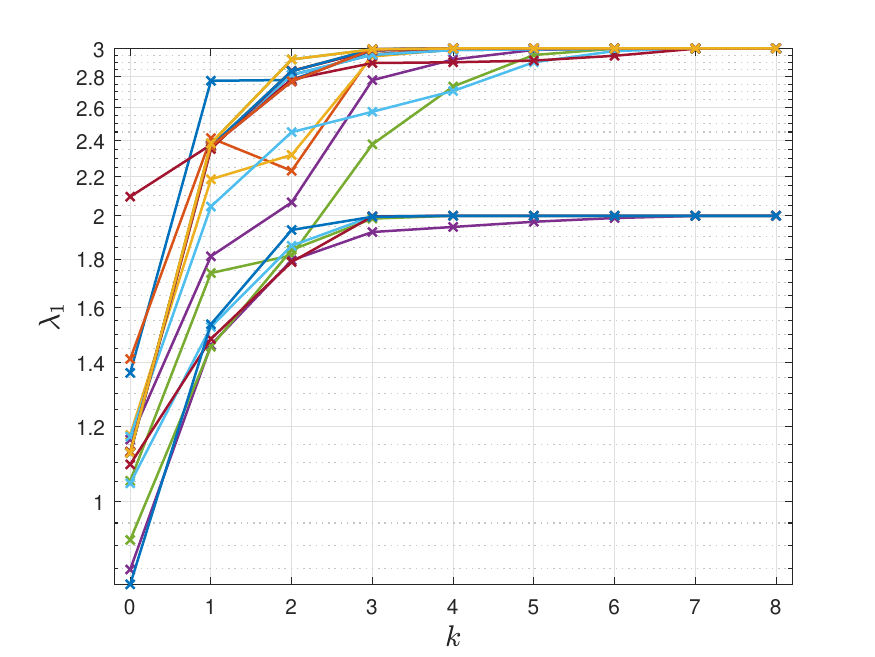}
    \caption{The largest eigenvalues during the rank minimization process while solving the problem in Fig. \ref{fig:quad1} is increased to the fixed values of traces, i.e., $\trace(\mY_i)=3$ and $\trace(\mY_{\tau i})=2$, indicating rank-1 solutions.}
    \label{fig:sawyer_quads_rankmin}
\end{figure}

Fig. \ref{fig:sawyer_quads} shows two postures solved for an example where five quadrotors hover in front of the manipulator and the objectives are set differently: one with levelness cost alone and the other with a combination of levelness and centering (function \eqref{eq:concentrate}) costs. As seen in Fig. \ref{fig:sawyer_quads}, the quadrotors are restricted in the circular cone mentioned in Section \ref{sec:perspective proj}. As expected, both of the images are level with the ground while the quadrotors in the right image are closer to the center. Fig. \ref{fig:sawyer_quads_rankmin} depicts the computational process of the solution in Fig. \ref{fig:quad1} by showing the changes of the largest eigenvalues, $\lambda_1$, which as seen, is increased to the maximal values fixed by the traces $\trace(\mY_i)=3$ and $\trace(\mY_{\tau i})=2$.

To test the performance of our method on different problems, we first construct two sets of objects uniformly distributed in spaces $\{\mT_i\}_{i\in\cV_o}$ of two different sizes, marked as ``condensed'': $\Mat{2.2&-1.5&-0.5}^\trans\leq\mT_i\leq\Mat{6.8&1.5&6.5}^\trans$ and ``scattered'': $\Mat{1.4&-2&-1}^\trans\leq\mT_i\leq\Mat{7.6&2&7}^\trans$. Then, the generated sets of points are used as inputs in Algorithm \ref{alg:vik} with different vision-based costs. The simulation is performed on Intel Core i7-10510U at 2.30 GHz CPUs with MOSEK \cite{mosek} as solver. In this simulation, we use a quaternion parameterization introduced in \cite[section VIII]{wu2024tac_archive} as an additional technique to reduce the number of variables. It is shown in \cite{wu2024tac_archive} that using this parameterization can improve the computation speed.
To test how the method minimize the reprojection error, we use solutions from tests with levelness plus centering costs to generate images. Then, we add random noise ($\leq10^{-2}$) to the images and match them using our method. The results are presented in Table \ref{tab:performance}, where each row shows the results obtained from 500 tests, including success rate and for successful solutions: averaged computation time spent on solving the SDPs, averaged iterations taken, averaged cost increase $\Delta \bar{f}$ during the rank minimization algorithm, maximal distance of the rotation matrices $\{\mR_i\}$ from their projections on $\SO{3}$, $\{P(\mR_i)\}$, and the maximal second largest eigenvalues of $\mY_i$ and $\mY_{\tau i}$. 

It is seen in Table \ref{tab:performance} that the method can recover matrices that are very close to real rotation matrices and can minimize rank to be very close to 1. In some simulations, the solver fails to find a solution within the loop in steps \ref{step:startwhile2}-\ref{step:endwhile2} in limited attempts $p_{max}$. This is because the algorithm guarantees only local convergence, and if started from a bad initial point, the algorithm can get stuck at solutions with rank higher than one. It is also observed that adding more points and expanding the distribution can reduce the success rate and slow down the computation, i.e., making the problem more difficult to solve. In general, the results in Table \ref{tab:performance} show that the proposed method can find solutions to visual inverse kinematics problems that have various costs and inputs.

\begin{table*}[h]
    \centering
    \begin{tabular}{cccccccc}
    \toprule
        Distribution & Number of points & Type of cost & Avg. time / iterations & Success rate & $\Delta \bar{f}$ &$\max(\norm{\mR_{i}-P(\mR_{i})}_F)$ & maximal $e_2$\\
        \midrule
        Condensed & 5 & \circled{1} & $1.4892$(s)/$8.1820$ & $100\%$ & $0.1429$ &$2.7924\cdot 10^{-4}$ &$4.1864\cdot 10^{-5}$\\ 
        \midrule
        Condensed & 5 & \circled{1}+\mymk{2a} & $1.3432$(s)/$7.8380$ & $100\%$ & $0.2749$ &$2.7250\cdot 10^{-4}$ &$3.2707\cdot 10^{-5}$\\
        \midrule
         Condensed & 5 & \circled{1}+\mymk{2b} & $1.5849$(s)/$6.4334$ & $94.6\%$ & $0.3907$ &$2.7667\cdot 10^{-4}$ &$3.6456\cdot 10^{-5}$\\
         \midrule
         Condensed & 15 & \circled{1}+\mymk{2a} & $4.6871$(s)/$8.5991$ & $91.8\%$ & $0.4491$ &$2.7834\cdot 10^{-4}$ &$2.3803\cdot 10^{-5}$\\
         \midrule
        Condensed & 15 & \circled{3} & $2.2349$(s)/$7.0331$ &$98.7\%$ & $0.0178$ & $2.7050\cdot10^{-4}$&$4.0702\cdot 10^{-5}$\\
        \midrule
        Scattered & 5 & \circled{1} & $2.5394$(s)/$9.2511$ & $87.6\%$ &$0.1770$ &$2.7953\cdot10^{-4}$& $6.4208\cdot 10^{-5}$\\
        \midrule
        Scattered & 5 & \circled{1}+\mymk{2a} &$1.6593$(s)/$8.4910$ & $88.8\%$ &$0.3738$ &$2.7873\cdot 10^{-4}$&$2.7337\cdot 10^{-5}$\\
        \midrule
        Scattered & 15 & \circled{1}+\mymk{2a} &$5.6191$(s)/$9.023$ & $69.6\%$ &$0.7133$ &$2.8161\cdot 10^{-4}$&$3.3635\cdot 10^{-5}$\\
        \midrule
        Scattered & 15 & \circled{3} & $2.3684$(s)/$7.0708$ &$97.4\%$ & $0.0060$ & $2.5488\cdot10^{-4}$&$7.2774\cdot 10^{-6}$\\
         \bottomrule
    \end{tabular}
    \footnotesize{Types of cost are labeled as: \circled{1} levelness; \mymk{2a} centering using \eqref{eq:concentrate}; \mymk{2b} centering using \eqref{eq:concentrate_close}; \circled{3} minimizing the reprojection error.}
    \caption{Performance of the proposed method on different visual inputs and vision-based costs. Each row shows results from 500 tests.}
    \label{tab:performance}
\end{table*}

\section{Conclusions}
In this work, we introduce the visual inverse kinematics problem and formulate the camera visibility constraint as SDP constraints on a series of virtual prismatic joints. We provide multiple vision based costs to fulfill different objectives. We then provide a way to find solutions using a semidefinite relaxation followed by a rank minimization technique on fixed-trace matrices.





\bibliographystyle{IEEEtran} 
\bibliography{main}

\end{document}